\pdfoutput=1

\documentclass[11pt]{article}

\PassOptionsToPackage{table,xcdraw}{xcolor}
\usepackage[final]{acl}

\usepackage{times}
\usepackage{latexsym}

\usepackage[T1]{fontenc}

\usepackage[utf8]{inputenc}

\usepackage{microtype}

\usepackage{inconsolata}

\usepackage{graphicx}

\usepackage{algorithm}
\usepackage{algorithmic}
\usepackage{booktabs}
\usepackage{amsmath,amssymb,amsfonts}
\usepackage{bbding}
\usepackage{epigraph}
\usepackage{multirow}
\usepackage{graphicx}
\usepackage{pifont}
\usepackage{arydshln}
\usepackage{cleveref}
\usepackage{xspace}

\newcommand\venustitlefont[1]{{\fontfamily{cmtt}\selectfont\textbf{#1}}}
\newcommand{\DATASETNAME}{\venustitlefont{VENUS}\xspace}
\newcommand{\MODELNAME}{\venustitlefont{MARS}\xspace}

\newcommand{\myxmark}{\ding{55}}
\newcommand{\mychecking}{\ding{51}}

\title{Speaking Beyond Language: A Large-Scale Multimodal Dataset for Learning Nonverbal Cues from Video-Grounded Dialogues}

\author{
 \textbf{Youngmin Kim\textsuperscript{$\spadesuit$}\thanks{Equal contribution.}} \quad
 \textbf{Jiwan Chung\textsuperscript{$\spadesuit*$}} \quad
 \textbf{Jisoo Kim\textsuperscript{$\spadesuit$}} \quad
 \textbf{Sunghyun Lee\textsuperscript{$\spadesuit$}} \quad
\\
 \textbf{Sangkyu Lee\textsuperscript{$\spadesuit$}} \quad
 \textbf{Junhyeok Kim\textsuperscript{$\spadesuit$}} \quad
 \textbf{Cheoljong Yang\textsuperscript{$\clubsuit$}} \quad
 \textbf{Youngjae Yu \textsuperscript{$\spadesuit$}}
\\
 \textsuperscript{$\spadesuit$} Yonsei University \quad
 \textsuperscript{$\clubsuit$} NC Research, NCSOFT Corporation
\\
\\
\texttt{winston1214@yonsei.ac.kr}
}

\begin{document}
\maketitle
\begin{abstract}
Nonverbal communication is integral to human interaction, with gestures, facial expressions, and body language conveying critical aspects of intent and emotion. However, existing large language models (LLMs) fail to effectively incorporate these nonverbal elements, limiting their capacity to create fully immersive conversational experiences. We introduce \MODELNAME, a multimodal language model designed to understand and generate nonverbal cues alongside text, bridging this gap in conversational AI.
Our key innovation is \DATASETNAME, a large-scale dataset comprising annotated videos with time-aligned text, facial expressions, and body language.
Leveraging \DATASETNAME, we train \MODELNAME with a next-token prediction objective, combining text with vector-quantized nonverbal representations to achieve multimodal understanding and generation within a unified framework.
Based on various analyses of the \DATASETNAME datasets, we validate its substantial scale and high effectiveness. Our quantitative and qualitative results demonstrate that \MODELNAME successfully generates text and nonverbal languages, corresponding to conversational input.
Our dataset and code are available at~\href{https://github.com/winston1214/nonverbal-conversation}{https://github.com/winston1214/nonverbal-conversation}.
\end{abstract}

\section{Introduction}
\label{sec:intro}


Human conversations are a complex interplay of verbal and nonverbal-cues. Beyond spoken words, facial expressions, gestures, and body language play an integral role in conveying emotions, intentions, and subtle meanings~\cite{phutela2015:importance}. For instance, “Do you know what time it is?” with a neutral expression seeks information, while a frown and crossed arms imply a rebuke. These nonverbal elements are essential for creating rich and nuanced interactions.


Recent advancements in large language models (LLMs) have resulted in conversational agents that closely resemble human interactions in written form. However, these models are still predominantly limited to text-based communication, overlooking the crucial role of nonverbal expressions. Although recent works~\cite{ng2022:learning, park2024:let} have made strides in addressing this gap, they have primarily concentrated on facial expressions, neglecting the broader spectrum of body language, which is essential for more realistic and immersive communication.

A major challenge in developing multimodal conversational agents lies in the lack of large-scale training datasets. Existing video conversation datasets are either limited in scale or lack annotated nonverbal cues, as summarized in \cref{tab:comp_dataset}. To address this, we introduce \textbf{\DATASETNAME} (\textbf{V}id\textbf{E}o with \textbf{N}onverbal cues and \textbf{U}tterance \textbf{S}et), a novel corpus designed for multimodal conversations with nonverbal annotations.
\DATASETNAME consists of 10-minute clips from dialogue-rich podcasts featuring two-person interactions, carefully curated to ensure accurate speaker diarization and motion tracking. Transcriptions were generated using Speech-to-Text (STT) models, while pseudo-3D motion parameters were extracted and annotated separately for facial expressions and body gestures, providing a detailed resource for aligning verbal and nonverbal cues.


Using \DATASETNAME, we develop \textbf{\MODELNAME},~\textbf{M}ultimodal l\textbf{A}nguage Model with nonve\textbf{R}bal-cue\textbf{S}, a multimodal conversational agent capable of understanding and generating nonverbal cues alongside textual context in dialogues. Nonverbal cues, such as facial expressions and body movements, are represented as discrete latent tokens, compressed using VQ-VAE~\cite{van2017:neural}. Both textual and nonverbal tokens are trained jointly with a unified next-token prediction objective, enabling natural modeling of multimodal dialogues within a single framework.


We conduct extensive quantitative and qualitative analyses to evaluate the contributions of \DATASETNAME and \MODELNAME to multimodal dialogue modeling. First, we examine the distributional diversity of nonverbal elements in \DATASETNAME (\cref{sec:data_analysis}). Next, we assess the trade-off between compression efficiency and reconstruction quality of nonverbal token discretizers in \cref{subsec:quantization}. Finally, we evaluate the multimodal conversational modeling capabilities of the \MODELNAME LLM in \cref{subsec:eval_mllm}.

Our key contributions are as follows:
\begin{itemize}
    \item Introduction of \DATASETNAME, the first large-scale multimodal conversational dataset designed for modeling nonverbal expressions.
    \item Development of \MODELNAME, a multimodal conversational agent leveraging \DATASETNAME to enable both the understanding and generation of nonverbal expressions within dialogue contexts.
    \item Comprehensive experimental validation, demonstrating the effectiveness of multimodal tokens in \MODELNAME for producing natural and contextually aligned nonverbal expressions alongside text, supported by user studies, quantitative evaluations, and qualitative analyses.
\end{itemize}

\begin{figure*}[ht!]
    \centering
    \includegraphics[width=\linewidth]{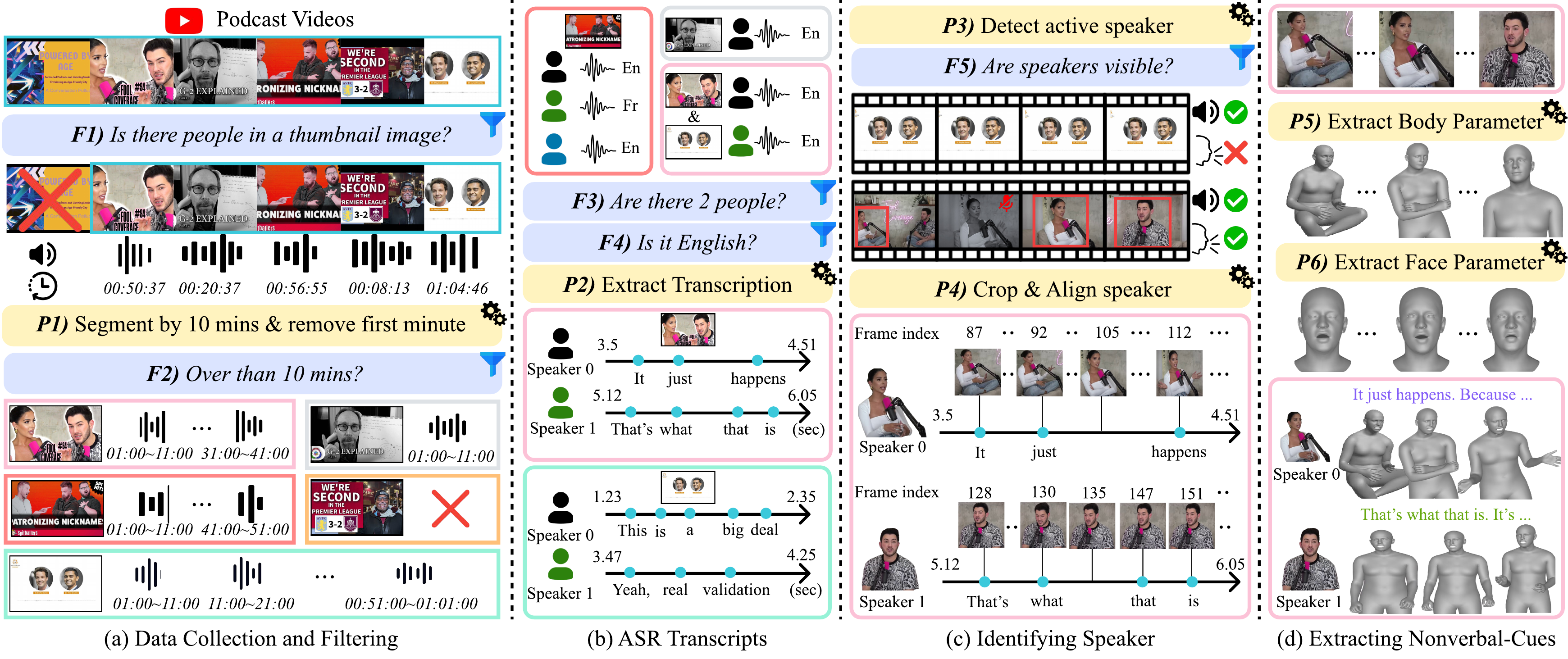}
    \caption{\textbf{Overview of \DATASETNAME collection pipeline.} (a) and (b) use only audio information, while (c) and (d) also utilize visual information. The blue boxes contain filtering criteria (\textit{F}), and the yellow boxes pertain to the processing steps (\textit{P}). The final box shown in (d) represents the facial expression and body language combined and represented using SMPL-X parameters. For more details, refer to the Section~\ref{sec:VENUS}.}
    \label{fig:data_collection}
\end{figure*}
\section{Related Works}
\label{sec:related_works}
\noindent \textbf{Multimodal Large Language Models.}
Recent studies have introduced models that combine various modalities with large language models (LLMs), extending their capabilities beyond text to include visual, auditory, and multimodal reasoning. Specifically, to enhance visual comprehension capabilities of LLMs, LLaVA~\cite{liu2024:visual}, Qwen-VL~\cite{bai2023:qwen} and MiniGPT-4~\cite{chen2023:sharegpt4v} have successfully integrated vision encoders into pre-trained LLMs.
Furthermore, VideoChat~\cite{li2023:videochat} and Video-LLaMA~\cite{zhang2023:video} extend these capabilities to video understanding, while models such as Unified-IO-2~\cite{lu2024:unified} and GPT-4-O~\cite{achiam2023:gpt} expand the scope to include auditory modalities, showing robust multimodal reasoning across various inputs.

\noindent \textbf{Learning Dialogue in Video.}
The importance of analyzing conversational sentiment using multimodal data (\textit{e.g.}, text, audio, and visual) from videos has driven the development of numerous datasets~\cite{busso2008:iemocap, zadeh2018:multimodal, poria2019:meld}. 
This has further spurred research into generating and understanding dialogues from videos, leveraging multimodal cues. For instance, Champagne~\cite{han2023:champagne} introduced the YTD-18M dataset for dialogue generation using visual signals and LLMs, while MultiDialog~\cite{park2024:let} combined audio and visual data for generating conversations. Beyond text, efforts like~\cite{shafique2023:nonverbal} and EmotionCLIP~\cite{zhang2023:learning} focus on recognizing nonverbal cues, such as gestures and emotions. Additionally, works like FurChat~\cite{cherakara2023:furchat} and~\cite{lee2023:developing} explore applying nonverbal signals to enhance robotic facial expressions and actions.
However, existing conversational datasets are often limited in scale or fail to include detailed 3D facial and body language information necessary for modeling nonverbal cues effectively. Our VENUS dataset addresses these gaps by being both large-scale and scalable, offering comprehensive conversational data that integrates not only text but also 3D facial expressions and body languages. This enables a more nuanced understanding of nonverbal cues and supports the generation of richer, context-aware conversations.

\noindent \textbf{Human Motion Synthesis in Conversation.}
Recent advancements in 3D human reconstruction~\cite{lin2023:one, dwivedi2024:tokenhmr, danvevcek2022emoca} have significantly improved the quality of pseudo-ground truth data, providing a scalable and accessible alternative to traditional sensor-based methods~\cite{yi2023:generating}.
Leveraging these datasets, recent works~\cite{wu2024:motionllm, lu2023:humantomato} have focused on generating human motions from text.
Building on this progress, our work utilizes pseudo labels derived from our VENUS, which addresses the lack of large-scale dataset for conversational settings. Unlike previous works like~\cite{ng2023:can, ng2022:learning}, which primarily generate listener facial motions from text, our approach extends to produce text, facial expressions, and body language, aligned with conversational context. 
\section{Learning Real-World Conversation with Nonverbal-Cues}
Previous studies have primarily focused on dialogue models and datasets that consider either text alone or text along with facial expressions. 
However, real conversations rely on both facial expressions and body gestures, utilizing the whole body for effective communication. To address this gap, we propose a dialogue model, \MODELNAME, for realistic interactions. 
Since no existing dataset simultaneously aligns text, facial expressions, and body language, we constructed a large-scale dataset, \DATASETNAME, in which text, facial expressions, and body language are aligned in the wild.

\subsection{\DATASETNAME: Video with Nonverbal-Cues and Utterance Set}
\label{sec:VENUS}
In this section, we introduce our pipeline to collect \DATASETNAME, which is outlined in Figure~\ref{fig:data_collection}.
Further details can be found in Appendix~\ref{sec:app_detail_dataset}. 

\noindent \textbf{Data Collection and Filtering.}
We collected YouTube podcast videos to learn nonverbal expressions included in conversations. 
Our goal was to efficiently extract and collect extensive conversation data from YouTube videos with only two people conversing.
We followed the filtering process presented in~\cite{han2023:champagne, zellers2021:merlot}. 
Initially, we screened thumbnails using a lightweight detector model~\cite{yolov8} to check for the presence of people, discarding videos without any people in the thumbnails (\textbf{\textit{F1}}).
We then removed the first minute to eliminate opening music or other introductory content (\textbf{\textit{P1}}).
Subsequently, to maximize the extraction of information from each video, we segmented each video into $10$-minute segments and discarded any segments shorter than $10$ minutes (\textbf{\textit{P1 \& F2}}).
In this step, we set the frames per second (FPS) at $25$.

\noindent \textbf{Automatic Speech Recognition Transcripts.}
To train the conversational model, we collected videos featuring interactions between two speakers.
We only downloaded audio to collect and filter videos, which is a cost-effective strategy.
Using PyAnnote~\cite{bredin2020:pyannote}, we performed speech diarization to identify videos with precisely two speakers and discarded videos without exactly two speakers (\textbf{\textit{F3}}).

Next, we utilized the state-of-the-arts speech-to-text model, WhisperX~\cite{bain2023:whisperx}, to filter and retain only English videos (\textbf{\textit{F4}}).
For these selected videos, we leveraged WhisperX to generate time-aligned speech transcripts (\textbf{\textit{P2}}).
By aligning the results predicted by the two models, we extracted the speaker's transcript at the word, sequence, and utterance levels.

\noindent \textbf{Identifying Speakers in Video.}
To effectively extract verbal and nonverbal features from videos, it is crucial to distinguish between the speaker and the listener.
To achieve this, we utilized the Light-ASD~\cite{liao2023:light} active speaker detection model to identify speakers within the video (\textbf{\textit{P3}}).
Additionally, we integrated a pretrained person detector model~\cite{yolov8} to extract visual features associated with each speaker. 
Here, we can extract frames with the speaker and their bounding box coordinates.
If the number of predicted speaker frames is less than the more number of predicted words from WhisperX, we consider it to lack visual variation and discard it (\textbf{\textit{F5}}).
Then, we cropped the speaker's image, $f$, using the detected speaker's bounding boxes.
To handle cases where multiple speakers are speaking simultaneously, we used a lightweight model~\cite{sandler2018:mobilenetv2} to extract the features of each speaker and align the speaker's images by comparing them with previous frames based on cosine similarity (\textbf{\textit{P4}}). The specific steps of this process are detailed in the Appendix~\ref{sec:app_reannote}.

To align the text and the speaker's frames, we segmented the speech into utterances in a video.
Then, using the time and FPS of the speaker's video, we calculate the set of frames for each utterance, $U_j=\{f_1, f_2, \cdots, f_i\}$.
Through this calculation, we can construct a set of $u$ utterances, $\mathcal{U} = [U]^{u}_{j=1}$, for each video.

\noindent \textbf{Extracting Nonverbal-Cues.}
We represent nonverbal cues as 3D parameters and, following the previous approaches~\cite{lin2024:motion, liu2024:emage}, extract facial parameters using the FLAME~\cite{li2017:learning} and body and hand gesture parameters using the SMPL-X~\cite{pavlakos2019:expressive}. 
To achieve this, we used EMOCA-v2~\cite{lu2023:audio} for facial expression and OSX~\cite{lin2023:one} for the whole body, extracting the parameters $M^{f}_{j} = \{m^{f}_{l}\}^{|U_j|}_{l=1} \ \text{where, } m^{f}_{l} \in \mathbb{R}^{156}$ and $M^{b}_{j} = \{m^{b}_{l}\}^{|U_j|}_{l=1} \ \text{where, } m^{b}_{l} \in \mathbb{R}^{179}$, respectively (\textbf{\textit{P5} \& \textit{P6}}). 
Finally, we annotated the video with nonverbal expressions, represented as 3D parameters that are aligned with the text for each utterance.

\begin{figure*}[t!]
    \centering
    \includegraphics[width=\linewidth]{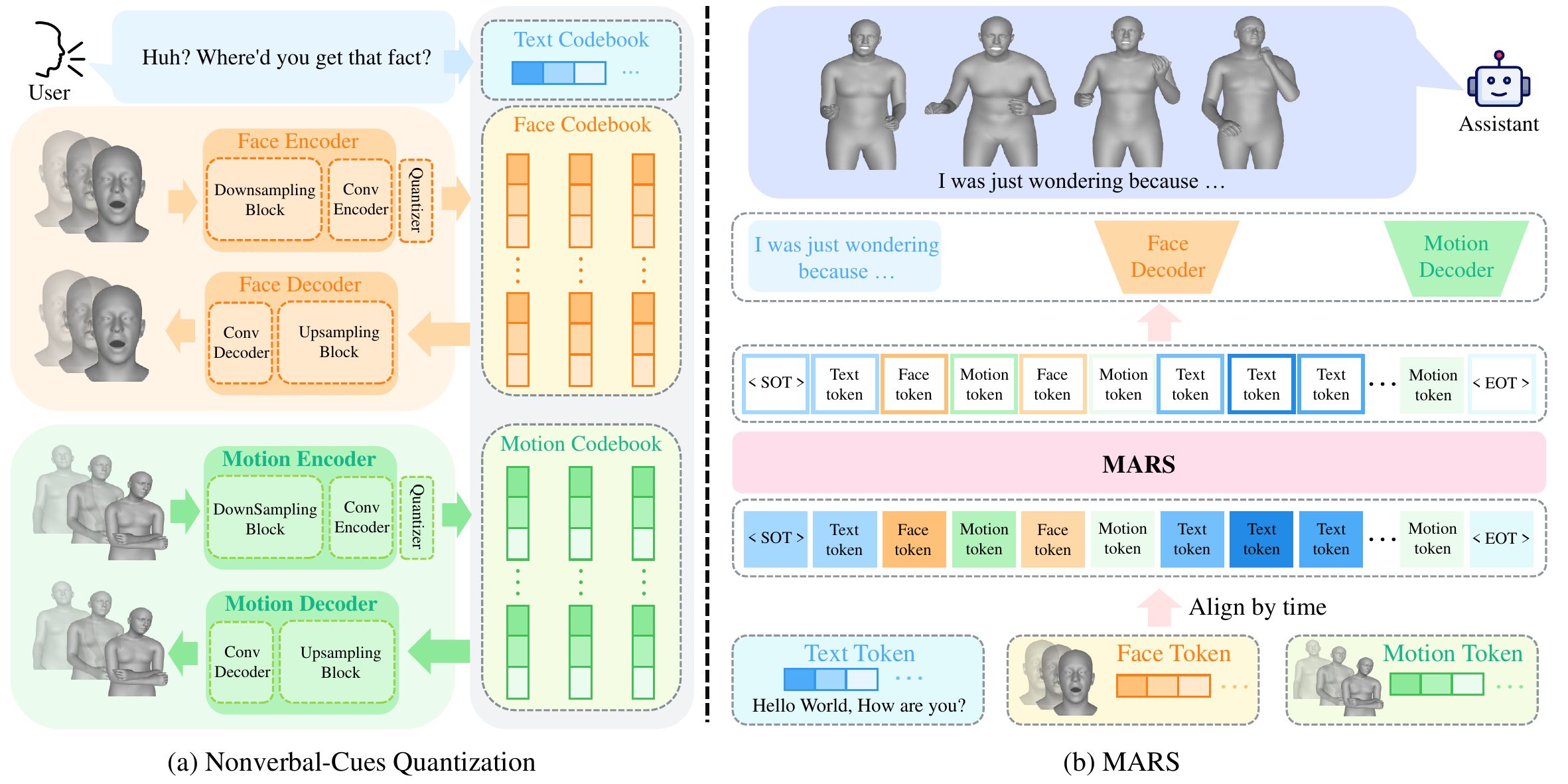}
    \caption{\textbf{System overview}. Our system consists of two main parts: (a) the VQ-VAE model trained to quantize nonverbal cues, and (b) a \MODELNAME trained to process quantized nonverbal expressions alongside text. The output generated by the assistant is visualized by replacing both face and body parameters with SMPL-X.}
    \label{fig:model}
\end{figure*}

\begingroup
\setlength{\dashlinegap}{1.5pt}
\setlength{\arrayrulewidth}{0.3pt}
\begin{table*}[t]
    \centering

    \resizebox{\linewidth}{!}{
    \begin{tabular}{c|cccccc}
    \toprule
        \textbf{Dataset} & \textbf{\# Dialogues} & \textbf{\# Turns} & \textbf{Length (hrs)}  & \textbf{Text} & \textbf{Video} &  \textbf{Nonverbal cues} \\ \toprule
        IEMOCAP~\cite{busso2008:iemocap}  & $151$ & $7,333$ & $12$ & \mychecking & \mychecking & \myxmark \\ 
        CMU-MOSEI~\cite{zadeh2018:multimodal}  & $3,228$ & - & $65$ & \mychecking & \mychecking & \myxmark  \\
        MELD~\cite{poria2019:meld}  & $1,433$ & $13,708$ & $13.7$ & \mychecking & \mychecking & \myxmark  \\ 
        YTD-18M~\cite{han2023:champagne}  & $\mathbf{18}$M & $\mathbf{54}$M${}^{*}$ & $\mathbf{30}$K${}^{*}$ & \mychecking & \mychecking & \myxmark  \\ 
        MultiDialog~\cite{park2024:let}  & $8,733$ & $187,859$ & $340$ & \mychecking & \mychecking & \myxmark \\ 
        \hdashline 
        BEAT~\cite{liu2022:beat} & \myxmark & \myxmark & $76$ & \mychecking & \myxmark & \mychecking \\
        EMAGE~\cite{liu2024:emage} & \myxmark & \myxmark & $60$ & \mychecking & \myxmark & \mychecking \\
        TalkShow~\cite{yi2023:generating} & \myxmark & \myxmark & $27$ & \myxmark & \myxmark & \mychecking \\ \midrule
        Ours (VENUS) & $\underline{89,459}$ & $\underline{1,114,328}$ & $\underline{14,910}$ & \mychecking & \mychecking & \mychecking  \\ 
        \bottomrule
    \end{tabular}
    }
    \caption{\textbf{Comparison of the \DATASETNAME dataset with the previous conversational and 3D gesture dataset.} The first block represents the conversation dataset, while the second block represents the gesture dataset. ``*'' represents an estimated value. For \textbf{\# Turns}, it was calculated by multiplying the average number of utterances per video $3$ by the number of videos. The~\textbf{Length (hrs)} was considered to be a maximum of 1 minute per video for the calculations. \textbf{Nonverbal cues} indicate whether 3D data or any other annotations for facial expressions or body language are provided. \textbf{Best} and \underline{second} are highlighted. Our dataset is the largest conversational dataset with annotations of nonverbal cues.}
    \label{tab:comp_dataset}
\end{table*}
\endgroup

\subsection{Nonverbal-Cues Quantization}
\label{sec:vqvae}
In this section, we introduce the tokenization process for large-scale collected nonverbal expressions from \DATASETNAME, as illustrated in Figure~\ref{fig:model}-(a).

\noindent \textbf{Notation and Problem Setup. }
We denote the sequence parameters of face and body movement at the utterance level as $M^{f}_{j} = \{m^{f}_{l}\}^{|U_j|}_{l=1}$ and $M^{b}_{j} = \{m^{b}_{l}\}^{|U_j|}_{l=1}$, respectively. 
We represent the facial components using the expression ($\psi$) and jaw parameters ($\theta^{jaw}$), resulting in $|\psi| + |\theta^{jaw}| = 53$ dimensions per frame (i.e., 50 expression parameters and 3 jaw pose parameters).
Similarly, for body language, we focus on the upper body ($\theta^{ubody}$), and the left and right hands ($\theta^{lhand}, \theta^{rhand}$)
This representation results in $|\theta^{ubody}| + |\theta^{rhand}| + |\theta^{lhand}|= 117$ dimensions per frame (i.e., 27 upper body parameters and 45 left and right hand parameters, respectively).
These are expressed as a sequence of $W$ frames, and to ensure smoothness, we apply the Savitzky–Golay method~\cite{gorry1990:general} to the sequence.
Therefore, the sequence of face and body parameters follows:
\begin{equation}
    \hat{M}^{f}_{j} = \{\hat{m}^{f}_l\}^{W}_{l=1} \quad \hat{M}^{b}_{j} = \{\hat{m}^{b}_l\}^{W}_{l=1},
\end{equation}
where $\hat{m}^{f}_{l} = [\psi_{l}, \theta^{jaw}_{l}] \in \mathbb{R}^{W \times 53}$ and $\hat{m}^{b}_{l} = [\theta^{ubody}, \theta^{rhand}, \theta^{lhand}] \in \mathbb{R}^{W \times 117}.$

\noindent \textbf{Architecture. }
To enable the conversational model, specifically the LLM, to understand nonverbal cues, we need to quantize continuous nonverbal features into discrete tokens.
To discrete tokenize nonverbal-cues, we adopted the architecture based on VQ-VAE~\cite{van2017:neural, razavi2019:generating}, which consists of an encoder-quantizer-decoder framework, to achieve this tokenization of nonverbal cues. For the purposes of this explanation, we will denote both input values $\hat{m}^{f}_{l}$ and $\hat{m}^{b}_{l}$ as $m_{l} \in \mathbb{R}^{W \times d}$ where $d$ is the length of the parameters, which can be either $53$ or $117$.

In this framework, the encoder, $E$, and decoder, $D$, are convolution networks with downsample ratio $q$, the quantizer contains a codebook $\mathcal{Z}\in \mathbb{R}^{K \times C}$, where $K$ denotes the codebook size and $C$ represents codebook dimension.
In the encoder process, when the sequence vector $m_{1:W}$ is input, it is downsampled to obtain latent vector $\mathbf{z}$, which follows:
\begin{equation}
    E(m_{1:W}) \rightarrow \mathbf{z} \in \mathbb{R}^{C \times \tau} \quad \text{where, } \tau = \frac{W}{q}.
\end{equation}
Given the latent vector $\mathbf{z}$ and the quantizer $\mathcal{Q}(\cdot ; \mathcal{Z})$, the quantized vector $\hat{\mathbf{z}}$ is determined as:
\begin{equation}
    \mathbf{\hat{z}} = \mathcal{Q}(\mathbf{z} ; \mathcal{Z}) = \arg\min_{e_k} \|\mathbf{z} - e_k\|_2^2,
\end{equation}
where $e_{k}$ denotes the $k$-th embedding in the codebook $\mathcal{Z}$. 
To stabilize training, we employ exponential moving averages (EMA) based codebook updates following~\cite{zhang2023:generating, guo2024:momask}.
The quantized vector $\hat{\mathbf{z}}$ is the element selected from the codebook that minimizes the reconstruction error with respect to $\mathbf{z}$.
During decoder process, the quantized latent vector $\mathbf{\hat{z}}$ undergoes upsampling process to reconstruct the original input sequence vector $m_{1:W}$.
\begin{equation}
    D(\mathbf{\hat{z}}) \rightarrow \hat{m}_{1:W} \in \mathbb{R}^{d}.
\end{equation}

Based on this architecture, we developed models for facial and body language, designated as Face VQ-VAE and Body VQ-VAE, respectively.

\noindent \textbf{Training losses. } 
We train Face VQ-VAE and Body VQ-VAE with the following loss functions $\mathcal{L}_{face}$ and $\mathcal{L}_{body}$, respectively:
\begin{equation}
    \begin{gathered}  
    \mathcal{L}_{face} = \mathcal{L}_{vq} + \lambda^{f}_{recon} \mathcal{L}^{f}_{recon} + \lambda^{f}_{vel} \mathcal{L}^{f}_{vel} \\
    \mathcal{L}_{body} = \mathcal{L}_{vq} + \lambda^{b}_{recon} \mathcal{L}^{b}_{recon} + \lambda^{b}_{vel} \mathcal{L}^{b}_{vel}
    \end{gathered}
\end{equation}
For codebook learning, we use commitment loss, $\mathcal{L}_{vq}$, in the proposed~\cite{van2017:neural}. 
\begin{equation}
    \mathcal{L}_{vq} = \beta||\mathbf{z} - \text{sg}(\mathbf{\hat{z}})||^{2}_{2},
\end{equation}
where $\text{sg}(\cdot)$ is a stop gradient operation and $\beta$ is commitment loss weight.

First, we introduce $\mathcal{L}^{f}_{recon}$ for the training of \textbf{Face VQ-VAE}.
For training face features reconstruction, the expression components $\psi_l$ and jaw, $\theta^{jaw}_{l}$ are separated, and each part is calculated, respectively. It follows: 
\begin{equation}
\begin{split}
\mathcal{L}^{f}_{recon} = & \lambda^{\psi}_{recon}L_{1}(\psi_{l} , \hat{\psi}_{l}) \\
& + \lambda^{jaw}_{recon}L_{1}(\theta^{jaw}_{l} , \hat{\theta}^{jaw}_{l}).
\end{split}
\end{equation}
Next, to preserve the temporal continuity and natural dynamics of facial motion, we design a facial motion velocity loss, $\mathcal{L}^{f}_{vel}$, as follows:
\begin{equation}
\begin{split}
    \mathcal{L}^{f}_{vel} = & L_{1}(v(\psi_{l}), v(\hat{\psi}_{l}))
    \\
    & + \lambda_{\theta}L_{1}(v(\theta^{jaw}_{l}), v(\hat{\theta}^{jaw}_{l})).
\end{split}
\end{equation}
Here, the function $v(p)$ computes the temporal velocity of a sequence $p$ by taking the frame-wise difference:
\begin{equation}
    v(p_{l}) = p_{l+1} - p_{l}.
\end{equation}

Similarly, the training objectives for the \textbf{Body VQ-VAE}, $\mathcal{L}^{b}_{recon}$, is defined similarly to those used in the Face VQ-VAE model.
For motion reconstruction, each component is calculated separately as $\mathcal{L}^{b}_{recon} = \Sigma^{i}_{body} L_{1} (\theta^{i} - \hat{\theta}^{i} )$ where,  $body \in \{ubody, rhand, lhand\}$.

\subsection{\MODELNAME: Multimodal Language Model with Nonverbal-Cues}
Using the quantized codebooks from Face VQ-VAE and Body VQ-VAE, the generation of text and nonverbal-cues sequences relies on their respective decoders and quantized representations.
Previous studies typically follow an auto-regressive approach; however, this cannot be directly applied when utilizing two codebooks. Inspired by methods proposed in studies that involve multiple codebooks~\cite{lu2023:humantomato}, we propose \textbf{\MODELNAME}, a multimodal language model with nonverbal-cues, designed to predict hierarchical discrete codes that capture nonverbal cues effectively.
This is illustrated Figure~\ref{fig:model} - (b).

\noindent \textbf{Training. }
The \MODELNAME is designed with the Transformer~\cite{vaswani2017:attention} architecture, where the input consists of textual tokens paired with corresponding nonverbal tokens.
The code indices corresponding to the facial expression and body language parameter sequences, $\hat{M}^{f}_{j}$ and $\hat{M}^{b}_{j}$, are denoted as $\mathbf{X}^{f} = [\mathbf{x}^{f}_1, \mathbf{x}^{f}_2, \cdots \mathbf{x}^{f}_{W/q}]$ and $\mathbf{X}^{b} = [\mathbf{x}^{b}_1, \mathbf{x}^{b}_2, \cdots \mathbf{x}^{b}_{W/q}]$, respectively.
Thus, the input tokens are composed of three elements: the word tokens $\mathbf{X}^{w} = [\mathbf{x}^{w}_1, \mathbf{x}^{w}_2, \cdots, \mathbf{x}^{w}_l]$, along with the facial and body code indices, $\mathbf{X}^{f}$ and $\mathbf{X}^{b}$.

Given that we input and generate nonverbal-cues corresponding to each word, the input sequences, $T$, are organized to align with their respective timestamps. 
\begin{equation}
    T = \{ \mathbf{x} \mid \mathbf{x}_i \in \bigcup_{c} X^{c} , c \in \{w,f,b\} \},
\end{equation}
where the sequence is ordered as $T = [ \mathbf{x}^{w}_{1}, \mathbf{x}^{f}_{1}, \mathbf{x}^{b}_{1}, \mathbf{x}^{w}_{2}, \cdots ]$.


Therefore, the word, face, and body token code indices prediction can be formulated as an auto-regressive prediction problem:
\begin{equation}
\begin{split}
p(T) = & \prod_{j=1}^{l} p_{\theta}(\mathbf{x}^{w}_j \mid T_{<j}) \\
& \prod_{k=1}^{W/q} \left[ p_{\theta}(\mathbf{x}_k^f \mid T_{<k}) \cdot p_{\theta}(\mathbf{x}_k^b \mid T_{<k}) \right],
\end{split}
\end{equation}
where $\theta$ represents the trainable parameters of the model. In this formulation, the word tokens are predicted first, followed by the face and body token indices.

\begingroup
\begin{table}[ht]
    \centering
    \resizebox{\columnwidth}{!}{
    \begin{tabular}{l:c}
    \toprule
        Total number of collected channels & $869$ \\ 
        Total number of collected videos & $27,128$ \\ 
        Total number of collected nonverbal expressions & $1\text{B}$ \\
        Total number of dialogues & $89,459$ \\ 
        Total number of turns & $1,114,328$ \\
        Total number of sentences & ~ $7,118,654$\\ 
        Total of unique words & $527,270$ \\
        Average number of turns per dialogue & $21$ \\
        Average length of utterances per dialogue in words & $170.829$ \\
        Average length of utterances per dialogue in seconds & $55.305$ \\
        Average number of nonverbal expressions per utterance in frames & $547$ \\
        \bottomrule
    \end{tabular}
    }
    \caption{\textbf{Summary of \DATASETNAME statistics.} The ``video'' refers to the video before it is segmented into 10-minute intervals, while ``dialogues'' refers to the conversations extracted from the videos segmented into 10-minute intervals.}
    \label{tab:stat_dataset}
\end{table}
\endgroup
\section{VENUS Dataset Analysis}
\label{sec:data_analysis}
We conducted data analysis to demonstrate the quality of the \DATASETNAME dataset. Additional analysis results can be found in the Appendix~\ref{sec:app_detail_dataset}.

\begin{figure}[ht]
    \centering
    \includegraphics[width=\linewidth]{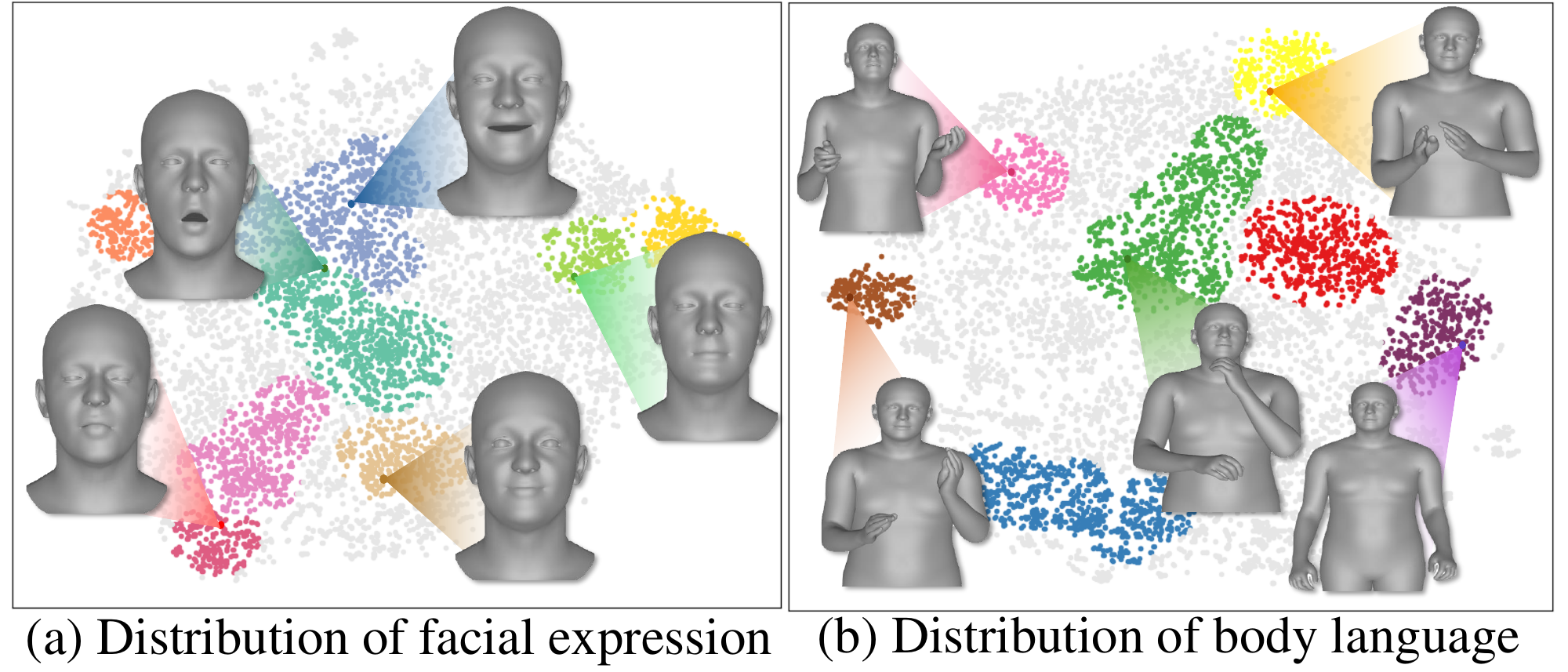}
    \caption{\textbf{Visualization of the distribution of nonverbal-cues}. (a) Facial expression embeddings are well-clustered despite the absence of emotion class labels, capturing meaningful emotion patterns. (b) Body language embeddings are similarly well-clustered, representing common conversational gestures that enhance communication or naturally occur during dialogue. Representative examples are provided for each cluster.}
    \label{fig:data_dist}
\end{figure}
\noindent \textbf{Statistic.}
The summary statistics of our dataset and comparison with statistics from other conversational and 3D gesture datasets are shown in Table~\ref{tab:stat_dataset} and Table~\ref{tab:comp_dataset}, respectively.
As shown in Table~\ref{tab:stat_dataset}, our dataset is large-scale, featuring lengthy utterances with numerous words and rich nonverbal expressions.
Each conversation averages 21 turns, which supports effective training for multi-turn dialogues.
Table~\ref{tab:comp_dataset} highlights that, compared to existing video-based multi-modal dialogue datasets, our dataset is the first to include annotations for nonverbal expressions.
While YTD-18M~\cite{han2023:champagne} has more videos, its conversations are segmented into intervals of up to one minute, potentially hindering context comprehension. In contrast, \DATASETNAME despite having fewer videos, includes longer conversations, making it better suited for understanding extended dialogues.
Furthermore, our dataset stands out as the largest-scale 3D annotated dataset when compared to previous 3D gesture datasets.


\begin{table*}[ht]
\resizebox{\textwidth}{!}{%
\begin{tabular}{@{}ccccccc|ccccc@{}}
\toprule
                                                      &     & \multicolumn{5}{c}{Face}                                          & \multicolumn{5}{c}{Body}                     \\ \cmidrule(lr){3-7} \cmidrule(lr){8-12} 
                                                      &     & VMSE ($10^{-1}$) $\downarrow$ & LVD ($10^{-3}$)$\downarrow$ & w-VL2 ($10^{-7}$) $\downarrow$ & Diversity $\uparrow$ & Variation $\uparrow$ & VMSE  $\downarrow$ & LVD ($10^{-1}$)$\downarrow$& w-VL2 ($10^{-4}$) $\downarrow$ & Diversity $\uparrow$ & Variation ($10^{-1}$) $\uparrow$ \\ \midrule
\multicolumn{2}{c}{GT}                                &  ~  &    ~    &  ~  &   $9.3323$  &   $0.8760$  & ~ &    ~   & ~ & $2.4189$ & $0.2803$  \\ \hdashline
\multicolumn{2}{c}{\cite{ng2023:can}}                 &    $0.5787$   & $0.4422$ &  $0.3832$  &$7.5866$&$0.5873$& $2.6424$ & $0.1268$ & $0.4338$ & $\mathbf{2.0151}$ & $\mathbf{0.1985}$  \\ 
\multicolumn{2}{c}{\cite{guo2024:momask}}             &    $0.5474$   & $0.4160$ &  $0.3429$  &   $7.7693$  &   $0.6253$  & $2.0608$ & $0.0994$ & $0.2100$ & $1.9934$ & $0.1951$  \\ 
\multicolumn{2}{c}{Ours}                              & \cellcolor[HTML]{EFEFEF}$\mathbf{0.5106}$ & \cellcolor[HTML]{EFEFEF}$\mathbf{0.4020}$ &\cellcolor[HTML]{EFEFEF}$\mathbf{0.2339}$ &\cellcolor[HTML]{EFEFEF}$\mathbf{7.8430}$ &\cellcolor[HTML]{EFEFEF}$\mathbf{0.6236}$   & \cellcolor[HTML]{EFEFEF}$\mathbf{1.9946}$ & \cellcolor[HTML]{EFEFEF}$\mathbf{0.0962}$ & \cellcolor[HTML]{EFEFEF}$\mathbf{0.2027}$ & \cellcolor[HTML]{EFEFEF}$1.9998$ & \cellcolor[HTML]{EFEFEF}$0.1956$ \\ \midrule               
\multirow{3}{*}{$L_{recon}$}                          & L1  & \cellcolor[HTML]{EFEFEF}$\mathbf{0.5106}$ & \cellcolor[HTML]{EFEFEF}$\mathbf{0.4020}$ & \cellcolor[HTML]{EFEFEF}$\mathbf{0.2339}$ & \cellcolor[HTML]{EFEFEF}$\mathbf{7.8430}$ & \cellcolor[HTML]{EFEFEF}$0.6236$ & \cellcolor[HTML]{EFEFEF}$\mathbf{1.9946}$ & \cellcolor[HTML]{EFEFEF}$\mathbf{0.0962}$ & \cellcolor[HTML]{EFEFEF}$\mathbf{0.2027}$ & \cellcolor[HTML]{EFEFEF}$\mathbf{1.9998}$ & \cellcolor[HTML]{EFEFEF}$0.1956$ \\
                                                      & L2  & $0.5471$ & $0.4124$ & $0.3630$ & $6.3334$ & $\mathbf{0.6425}$ & $2.3384$ & $0.1139$ & $0.3078$ & $1.9732$ & $0.1879$ \\ 
                                                      & smooth L1  & $0.4106$ &  $0.4034$    & $0.3313$  &  $6.3874$   &$0.6052$& $2.3210$ &  $0.1128$ & $0.2787$ &  $2.0603$ &  $\mathbf{0.2025}$  \\ \midrule

\multirow{6}{*}{Dim}                                  & 8   & \cellcolor[HTML]{EFEFEF}$\mathbf{0.5106}$ & \cellcolor[HTML]{EFEFEF}$\mathbf{0.4020}$ &\cellcolor[HTML]{EFEFEF}$0.2339$ &\cellcolor[HTML]{EFEFEF}$\mathbf{7.8430}$ &\cellcolor[HTML]{EFEFEF}$\mathbf{0.6236}$  & $2.0596$ &$0.0995$ &$0.2280$ & $1.9183$ & $0.1794$\\
                                                      & 16  & $0.5217$ & $0.4100$ & $0.2582$ & $7.6855$ & $0.6023$ & \cellcolor[HTML]{EFEFEF}$\mathbf{1.9946}$ & \cellcolor[HTML]{EFEFEF}$\mathbf{0.0962}$ & \cellcolor[HTML]{EFEFEF}$\mathbf{0.2027}$ & \cellcolor[HTML]{EFEFEF}$\mathbf{1.9998}$ & \cellcolor[HTML]{EFEFEF}$\mathbf{0.1956}$ \\
                                                      & 32  & $0.5294$ & $0.4150$ & $0.2439$ & $7.6986$ & $0.6006$ & $2.1199$ & $0.1022$ & $0.2192$ & $1.9838$ & $0.1926$ \\
                                                      & 64  & $0.5152$ & $0.4071$ & $0.2360$ & $7.6203$ &$0.5890$  & $2.1577$ &  $0.1037$ & $0.2312$ &  $1.9947$ & $0.1942$  \\ 
                                                      & 128 & $0.5222$ & $0.4153$ & $\mathbf{0.2314}$ & $7.7554$   & $0.6098$ & $2.1427$ & $0.1037$  & $0.2244$ & $1.9633$ & $0.1876$ \\
                                                      & 256 & $0.5296$ & $0.4183$ & $0.2443$ & $7.8247$ & $0.6212$ & $2.1410$ & $0.1034$ & $0.2387$ & $1.9936$ & $0.1939$ \\ \midrule
                                                      
\multirow{4}{*}{Size}                                 & 64  & $0.6628$ & $0.5181$ & $0.4472$ & $6.6604$ & $0.4566$ & $4.2495$ & $0.1993$ & $0.8084$ & $0.7093$ & $0.0306$ \\ 
                                                      & 128 & $0.5770$ & $0.4514$ & $0.3549$ & $7.3002$ & $0.5458$ & $2.1905$ & $0.1054$ & $0.2670$ & $1.9114$ & $0.1801$ \\
                                                      & 256 & $0.5313$ & $0.4184$ & $0.2583$ & $7.6053$ & $0.5890$ & $2.074$ & $0.1003$ & $0.2119$ & $1.9663$ & $0.1889$ \\
                                                      & 512 & \cellcolor[HTML]{EFEFEF}$\mathbf{0.5106}$ & \cellcolor[HTML]{EFEFEF}$\mathbf{0.4020}$ &\cellcolor[HTML]{EFEFEF}$\mathbf{0.2339}$ &\cellcolor[HTML]{EFEFEF}$\mathbf{7.8430}$ &\cellcolor[HTML]{EFEFEF}$\mathbf{0.6236}$  & \cellcolor[HTML]{EFEFEF}$\mathbf{1.9946}$ & \cellcolor[HTML]{EFEFEF}$\mathbf{0.0962}$ & \cellcolor[HTML]{EFEFEF}$\mathbf{0.2027}$ & \cellcolor[HTML]{EFEFEF}$\mathbf{1.9998}$ & \cellcolor[HTML]{EFEFEF}$\mathbf{0.1956}$\\ 
                                                      
\bottomrule
\end{tabular}%
}
\caption{\textbf{Experimental results on Face VQ-VAE and Body VQ-VAE.} ``$\mathcal{L}_{recon}$'' represents $\mathcal{L}^{f}_{recon}$ and $\mathcal{L}^{b}_{recon}$, ``Dim'' refers to the codebook embedding dimension, and ``size'' indicates the codebook size. Our key results are highlighted. The Face VQ-VAE achieved the best performance with L1 loss, an embedding dimension of $8$, and a codebook size of $512$, while the Body VQ-VAE performed best with L1 loss, an embedding dimension of $16$, and the same codebook size.}
\label{tab:vqvae_quan}
\end{table*}

\noindent \textbf{Distribution of Nonverbal Cues.}
To analyze the diversity of nonverbal expressions in our dataset, we sampled 10 random frames per video from approximately $1,000$ videos and applied T-SNE~\cite{van2008:visualizing} for dimensionality reduction. In Figure~\ref{fig:data_dist}, we display the results by creating 7 clusters for facial expressions and 8 clusters for body languages using DBSCAN~\cite{ester1996:density}.

Figure~\ref{fig:data_dist}-(a) displays the distribution of facial expressions, covering both the $\psi$ and $\theta^{jaw}$.  We can observe a variety of emotions, despite the absence of emotion labels. Notably, the blue and green points appeared the most since podcast conversations target to entertain or inform the viewers, leading to a larger portion of neutral and positive expressions.
In Figure~\ref{fig:data_dist}-(b) the distribution of body language $\theta^{ubody}$, $\theta^{lhand}$ and $\theta^{rhand}$ is displayed. The most common body language observed involves arms in a relaxed, lowered position, which typically reflects a conversational attitude. In addition, gestures that enhance or clarify the speaker's message, such as resting the chin on the hand or expressive hand movements, were frequently noted.

\section{Experiments}
\subsection{Experiment Setup}
We trained and evaluated our model using a subset of the VENUS dataset in our experiments
Both VQ-VAE and \MODELNAME were trained on $3,924$ videos and $69,412$ utterances.
For evaluation, VQ-VAE used the full test set consisting of $997$ videos and $30,390$ utterances, whereas \MODELNAME was evaluated on a subset of $1,000$ utterances sampled from the test set.

\subsection{Nonverbal-cues Quantization}
\label{subsec:quantization}
\noindent \textbf{Evaluation Metric. }
We quantitatively evaluate how realistically facial expressions and body languages have been quantized, based on evaluation methods proposed in previous studies~\cite{ng2022:learning, ng2023:can, liu2024:emage}. To this end, we adopt five metrics to assess the realism and diversity of facial expressions and body language.
To evaluate realism, we use \textbf{VMSE}, \textbf{LVD}, and \textbf{window Vertex L2}, while diversity is assessed using \textbf{diversity} and \textbf{variance}. Detailed explanations of these metrics are provided in the Appendix~\ref{appendix:vq_eval_metrics}.

\noindent \textbf{Results. }
We conducted an ablation study to evaluate our Face and Body VQ-VAE models, varying one component at a time (Table~\ref{tab:vqvae_quan}). Based on the results, we chose L1 loss for the Face VQ-VAE and L1 loss for the Body VQ-VAE, with embedding dimensions of 8 and 16, respectively. Both used a codebook size of 512. These settings outperformed previous works~\cite{ng2023:can, guo2024:momask}.

\begin{table}[t]
\resizebox{\linewidth}{!}{%
\begin{tabular}{ccccccc}
\toprule
& & & \multicolumn{2}{c}{Text} & \multicolumn{2}{c}{Nonverbal} \\
\cmidrule(lr){4-5} \cmidrule(lr){6-7}
~ & ~ & PPL $\downarrow$ & BERT $\uparrow$ & METEOR $\uparrow$ & NLL-F $\downarrow$ & NLL-B $\downarrow$ \\
\midrule
\multirow{2}{*}{LLaMA 1B} & zero-shot & $5427.1$ & $0.811$ & $0.110$ & $16.232$ & $17.039$ \\
& \MODELNAME       & $\mathbf{1665.8}$ & $\mathbf{0.834}$ & $\mathbf{0.130}$ & $\mathbf{8.676}$ & $\mathbf{5.330}$ \\
\midrule
\multirow{2}{*}{Qwen 1.5B} & zero-shot & $3315.5$ & $0.823$ & $\mathbf{0.116}$ & $15.019$ & $15.911$ \\
& \MODELNAME        & $\mathbf{2990.0}$ & $\mathbf{0.839}$ & $0.115$ & $\mathbf{8.812}$ & $\mathbf{6.144}$ \\
\midrule
\multirow{2}{*}{LLaMA 3B} & zero-shot & $5477.0$ & $0.818$ & $\mathbf{0.136}$ & $16.504$ & $17.574$ \\
& \MODELNAME        & $\mathbf{926.9}$ & $\mathbf{0.835}$ & $0.133$ & $\mathbf{8.057}$ & $\mathbf{5.325}$ \\
\midrule
\multirow{2}{*}{Qwen 3B} & zero-shot & $56781.1$ & $0.811$ & $\mathbf{0.131}$ & $20.850$ & $20.874$ \\
& \MODELNAME        & $\mathbf{800.0}$ & $\mathbf{0.839}$ & $0.123$ & $\mathbf{7.295}$ & $\mathbf{4.666}$ \\
\bottomrule
\end{tabular}%
}
\caption{\textbf{Quantitative results of \MODELNAME.} $\downarrow$ means a lower score is better, $\uparrow$ means a higher score is better. Here, ``NLL-F'' and ``NLL-B'' denote the negative log-likelihood (NLL) for face tokens and body tokens, respectively. \MODELNAME demonstrates superior precision in generating nonverbal cues, highlighting its effectiveness in producing both text and nonverbal expressions. }
\label{tab:quan_mars}
\end{table}

\subsection{Semantic Evaluation for \MODELNAME}
\label{subsec:eval_mllm}

\noindent \textbf{Training Settings.}
We employ LLaMA 3.2 Instruct~\cite{meta_ai_blog_llama3} and Qwen 2.5 Instruct~\cite{Yang2024:Qwen25TR} as the large language model.
To clarify the model’s role, we incorporated a system prompt that facilitates effective generation of both nonverbal and textual tokens.
Additionally, since the nonverbal token is added as a special token, we performed supervised fine-tuning to ensure model's understanding of them.
Further details can be found in the Appendix~\ref{sec:appendix_mars}.

\noindent \textbf{Evaluation metrics. }
To evaluate \MODELNAME, we separately assess the quality of its text and nonverbal token outputs, as ensuring accurate alignment between these token types is inherently challenging.
First, we use Perplexity (\textbf{PPL}) as a general measure for both text and nonverbal tokens. 
For text tokens, we use \textbf{BERT-score} and \textbf{METEOR} as evaluation metrics, while for nonverbal tokens, we rely on Negative log-likelihood (\textbf{NLL}).

\noindent \textbf{Quantitative Results. }
We compared the quantitative performance of the LLM~\cite{meta_ai_blog_llama3} and our \MODELNAME model. As shown in Table~\ref{tab:quan_mars}, the conventional LLM model showed limitations in understanding special tokens containing nonverbal information, failing to generate them properly. 
In contrast, \MODELNAME, which was trained by interleaving nonverbal tokens within the textual input, achieved the lowest perplexity and the highest BERTScore across all model sizes, indicating its superior ability to generate semantically coherent dialogues.
Furthermore, the significantly lower NLL scores for nonverbal cues demonstrate that \MODELNAME successfully captures and generates nonverbal behaviors.
These results not only validate the effectiveness of our approach in handling multimodal signals but also highlight the scalability of \MODELNAME, as its performance improves with larger model sizes in both textual and nonverbal generation tasks.

\begin{figure}[t]
    \centering
    \includegraphics[width=\linewidth]{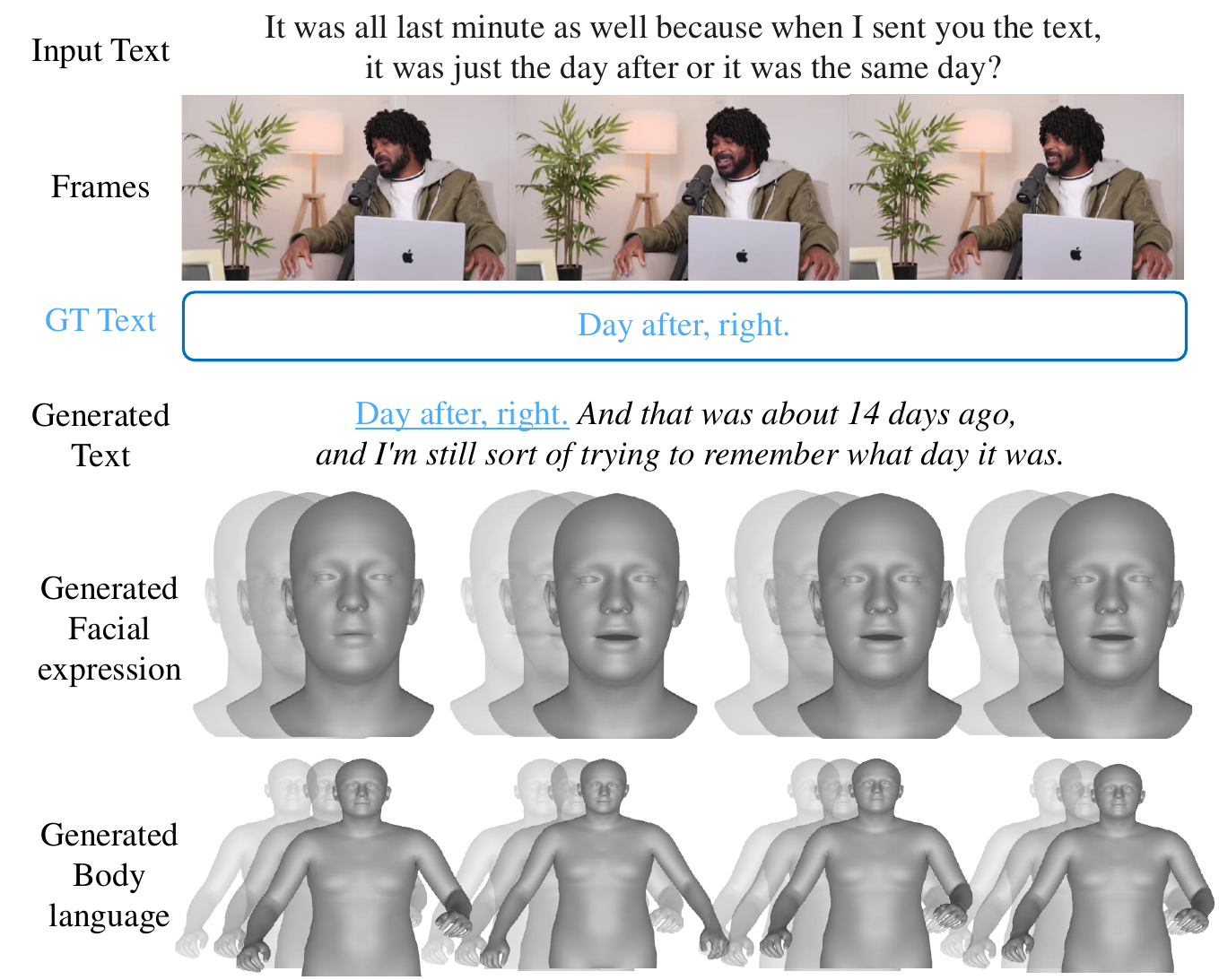}
    \caption{\textbf{Qualitative results for \MODELNAME.} Qualitative results showcasing inputs and outputs of our \MODELNAME model. Inputs include the user's text, face, and body language, while \MODELNAME outputs corresponding text, facial expressions, and body language. Underlined text indicates where \MODELNAME matches the ground truth (GT). Moreover, \MODELNAME produces improved text compared to GT and also successfully generates corresponding facial and body language aligned with the context.}
    \label{fig:mars_qual}
\end{figure}

\noindent \textbf{Qualitative Results. }
We use qualitative results to assess the effectiveness of our model in generating the listener's text and nonverbal expressions. As shown in Figure~\ref{fig:mars_qual}, our \MODELNAME not only aligns with the ground-truth (GT) but also produces more contextually enriched text and corresponding face and body languages. This demonstrates the qualitative effectiveness of our model in generating richer and more expressive listener responses.

\section{Conclusion}
In this work, we introduce \DATASETNAME, a video-based multimodal conversation dataset designed to understand and generate both text and nonverbal expressions, and present \MODELNAME. This language model can produce both dialogue and corresponding nonverbal behaviors. The \DATASETNAME dataset is built from YouTube videos, including real conversational text and the accompanying nonverbal cues (such as facial expressions and body language) annotated in 3D parameters. Using \DATASETNAME, our \MODELNAME model learns to align and generate both textual and nonverbal elements, resulting in more engaging and natural interactions. We believe that our \DATASETNAME dataset and \MODELNAME model will support a wide range of applications, such as virtual humans and gaming, by enabling the production of nonverbal behaviors in 3D.
\section{Limitations}
This study explores the development of a large language model (LLM) for generating nonverbal cues nameed \MODELNAME, supported by a custom dataset named \DATASETNAME designed to capture diverse nonverbal communication patterns. While the proposed approach demonstrates promising results, certain limitations remain that warrant further exploration.

First, the \DATASETNAME dataset utilized in this research is primarily curated from the Podcast channel, which may limit the diversity of nonverbal expression patterns in the data (e.g., crying or angry expressions). Furthermore, pseudo-labeling was employed in the dataset, which, while effective, could introduce potential inaccuracies that require further refinement. Additionally, not all data within the \DATASETNAME dataset was utilized, leaving room for broader exploration in future work.
Second, the evaluation metrics used in this study, though effective for assessing initial performance, may not fully capture the nonverbal communication. More sophisticated and comprehensive metrics are necessary to evaluate the system's performance in real-world scenarios.

Looking ahead, future work will aim to address these limitations by incorporating a wider range of nonverbal modalities, such as vocal expressions, to enrich the dataset and enhance the robustness of the model. Moreover, we plan to develop advanced evaluation metrics that better reflect the complexity of nonverbal communication. These improvements will further generalize and validate the applicability of our approach across diverse datasets and scenarios.

\section{Ethical Considerations}
In this paper, we introduce a large-scale multimodal conversational dataset named \DATASETNAME derived from publicly available YouTube videos. The dataset is designed to advance research in real-world conversational understanding by including frames, reconstructed facial expressions and body language of the interlocutors. While this dataset provides valuable insights for understanding conversational behavior, it may raise privacy concerns as it captures the visual and auditory cues of individuals.
To address these concerns, we follow ethical practices adopted by prior works~\cite{zellers2021merlot, zellers2022merlot, han2023:champagne} and release only the video IDs instead of the raw video frames. Additionally, the reconstructed face and body motions are represented as template meshes, ensuring anonymization and preventing direct identification of individuals.
To further protect user privacy, future directions may include further anonymizing faces and improving methods for deidentifying personal information. We remain committed to respecting user privacy and ensuring compliance with ethical standards in dataset creation and usage.

\section*{Acknowledgements}
This work was supported by NCSOFT, the National Research Foundation of Korea (NRF) grant funded by the Korea government (MSIT) (No. RS-2024-00354218), and the Institute of Information \& Communications Technology Planning \& Evaluation (IITP) grants funded by the Korea government (MSIT) (No. RS-2024-00457882, AI Research Hub Project; No. RS-2025-02263598, Development of Self-Evolving Embodied AGI Platform Technology through Real-World Experience).

\bibliography{custom}
\clearpage
\appendix

\section{Details of \DATASETNAME Dataset Collection}
\label{sec:app_detail_dataset}
In this section, we provide more details about \DATASETNAME that are not included in the main paper.
\subsection{Safety Filtering}
We utilized WildGuard~\cite{han2024:wildguard} to filter unsafe contents in video transcriptions.
WildGuard assesses the risk level(``harmful'' or ``unharmful'') and the parsing error on a single-turn basis for both prompts and responses. 
To maintain conversational context while applying safety filtering, we transformed video transcriptions into single-turn segments using a sliding-window approach.
Our safety filtering strategies are as follows: 1) An utterance is flagged as harmful if it is identified as such when considering both the prompt and the corresponding response. 2) An utterance is also deemed harmful if it is classified as harmful independently, whether it appears as a prompt or as a response, within a single turn.
3) If the cumulative duration of harmful utterances within a video exceeds three minutes, the entire video is discarded to ensure safety compliance.
By implementing these measures, we ensure robust safety filtering while preserving as much video information as possible.

\subsection{Video Collection Strategy}
To collect videos centered on conversations, we first used the YouTube API~\footnote{\url{https://developers.google.com/youtube}} to collect channel IDs that include the word ``Podcast'' in their channel names. 
After identifying these channels, we retrieved up to 300 videos per channel that were created between January 1, 2015, and December 31, 2023.
Due to the inherent limitations of the YouTube API, duplicate videos were occasionally retrieved during this process. To ensure the quality of the dataset, we removed all duplicates, retaining only unique videos.

\subsection{Re-annotate Speaker}
\label{sec:app_reannote}
To align the text by the speaker with nonverbal expressions, we segmented the speech into individual utterances in a video, $\mathcal{U} = [U_j]^{n}_{j=1}$ where $n$ is the number of utterances in a video.
Next, we used the time of the utterances, $T = [(t^{\text{start}}_{j}, t^{\text{end}}_{j})]^{n}_{j=1}$, extracted from WhisperX and the FPS to calculate the start and end frames of each utterance.
Then, we cropped the speaker's image to focus on the segments where the speaker is actively speaking. 
To handle speaker alignment, we used a lightweight model~\cite{sandler2018:mobilenetv2} to extract the features of the speaker's cropped images and re-aligned them by comparing with previous frames based on cosine similarity.
This is shown in Algorithm~\ref{algorithm:crop}.
\begin{algorithm}[t!]
\caption{Cropping and Aligning Speaker}
\label{algorithm:crop}
\small
\textbf{Input}: Frames with the speaker, $\mathcal{F} = [f_{i}]_{i=1}^{m}$, speaker's bounding box coordinates, $B$, and utterance start and end time, $T$.

\textbf{Output}: Utterance frames set without duplicates, $U_j$
\begin{algorithmic}[1]
    \STATE $(s_{j}, e_{j}) \gets \lfloor (t^{\text{start}}_{j}, t^{\text{end}}_{j}) \times \text{FPS} \rfloor$
    \STATE $F_j \gets \mathcal{F}[s_{j} : e_{j}]$
    \STATE $U'_j \gets []$
    \FORALL{$f$ in $F_j$}
        \STATE $u'_{j,k} \gets f[x^{j}_{\text{top}}:x^{j}_{\text{bottom}}, y^{j}_{\text{top}}:y^{j}_{\text{bottom}}]$
        \STATE Append $u'_{j,k}$ to $U'_j$
    \ENDFOR
    \STATE $U_j \gets \{\}$
    \STATE $u_{prev} \gets \text{None}$
    \FOR {each cropped frame $u'_{j,k}$ in $U'_j$}
        \IF {$k = 2$}
            \STATE $e_{p} \gets \text{MobileNet}(u_{prev})$
            \STATE $e_{j,1} \gets \text{MobileNet}(u'_{j,1})$
            \STATE $e_{j,2} \gets \text{MobileNet}(u'_{j,2})$
            \STATE $sim \gets \arg\max(\text{cos}(e_{j,1}, e_{p}), \text{cos}(e_{j,2}, e_{p}))$
            \STATE $u_j \gets u'_{j,sim}$
        \ELSE
            \STATE $u_j \gets u'_{j,1}$
        \ENDIF
        \STATE Append $u_j$ to $U_j$
        \STATE $u_{prev} \gets u_j$
    \ENDFOR
    \STATE \textbf{return} $U_j$
\end{algorithmic}
\end{algorithm}

\subsection{Batching for Nonverbal Cue Annotation}
To efficiently extract 3D information from a large corpus of speaker images, batch processing is essential. 
However, since we detect and crop speakers from video frames using the detection model, the resulting images $I \in \mathbb{R}^{h \times w}$ inherently vary in dimensions due to differences in the bounding boxes, where $h$ and $w$ denote height and width of each image, respectively.

To address the challenge of variable image sizes and enable batch inference, we propose a resizing and padding strategy that preserves the aspect ratio of each speaker image while standardizing their dimensions. The main idea is to scale each image such that its longest side matches a predetermined size $S$, followed by padding to create a square image of dimensions $S \times S$. Firstly, we compute the scaling factor $s$ based on the original dimensions of the image:
\begin{equation}
    s = \frac{S}{\max(w, h)}
\end{equation}

This scaling factor ensures that the largest dimension of the image is resized to $S$, maintaining the asepct ratio. The image is then resized to new dimensions $h' = s \times h$ and $w' = s \times w$.

After resizing, we create a zero-initialized square image $I_{\text{pad}} \in \mathbb{R}^{S \times S}$, and resized image $I_{\text{resized}} \in \mathbb{R}^{h' \times w'}$ is then placed at the center of $I_{\text{pad}}$ to ensure spatial consistency and preserve central features of the speaker. The offsets for centering are calculated as :

\begin{equation}
\delta_{\text{h}} = \left\lfloor \frac{S - h'}{2} \right\rfloor, \quad
\delta_{\text{w}} = \left\lfloor \frac{S - w'}{2} \right\rfloor
\end{equation}

The padded image $I_{\text{pad}}$ is then defined as:
\begin{equation}
\begin{array}{l}
    I_{\text{pad}}(i,j) \\[0.5em]
    = 
    \begin{cases} 
        I_{\text{r}}(i-\delta_{\text{h}},j-\delta_{\text{w}}) & \text{if } 
        \begin{aligned}
            i &\in \left[\delta_{\text{h}}, \delta_{\text{h}} + h'\right) \\
            j &\in \left[\delta_{\text{w}}, \delta_{\text{w}} + w'\right)
        \end{aligned} \\
        0 & \text{otherwise.}
    \end{cases}
\end{array}
\end{equation}

This approach maintains the aspect ratio of the original images and ensures that all images have a uniform size, facilitating efficient batch processing.

\subsection{Topic analysis}
\begin{figure}[t!]
    \centering
    \includegraphics[width=\linewidth]{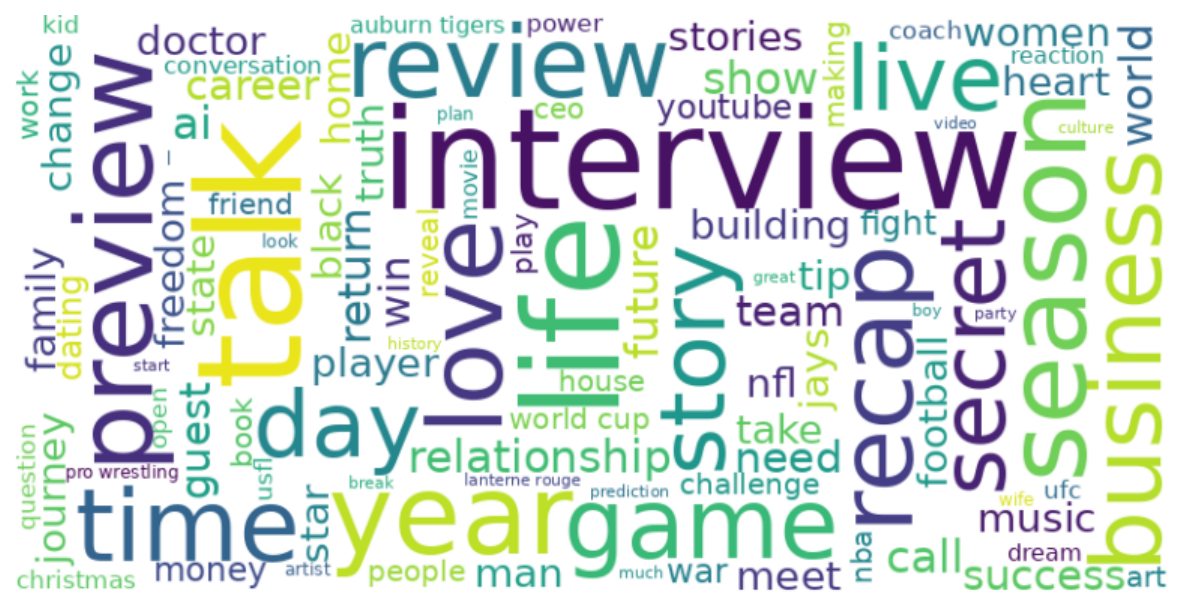}
    \caption{\textbf{The diversity of topics of videos in \DATASETNAME, displayed as a word cloud.} Larger words indicate more videos from that topic.}
    \label{fig:topic_wordcloud}
\end{figure}

We visualized the titles of videos from the entire dataset in Figure~\ref{fig:topic_wordcloud} as a Venn-style word cloud~\cite{coppersmith2014:dynamic}, with the size proportional to the number of videos gathered for that topic.
The most frequent 3 topics are interview ($6.64\%$), life ($4.51\%$), and recap ($4.3\%$).
As these proportions indicate, the topics of the \DATASETNAME videos are almost uniformly distributed, covering a wide range of conversational topics.

\subsection{Text-Based Sentiment Analysis}
For data analysis, we automatically predicted the sentiment (neutral, positive, negative) of the text using a Roberta-based sentiment classifier~\cite{camacho2022:tweetnlp}.
In the sentiment analysis conducted with \DATASETNAME at the sentence level, the results showed that $63.79\%$ of the sentences were classified as neutral, $17.36\%$ as positive, and $18.85\%$ as negative.
Based on the sentiment analysis results at the sentence level, we conducted a frequency analysis accordingly.

\begin{figure}[t!]
    \centering
    \includegraphics[width=\linewidth]{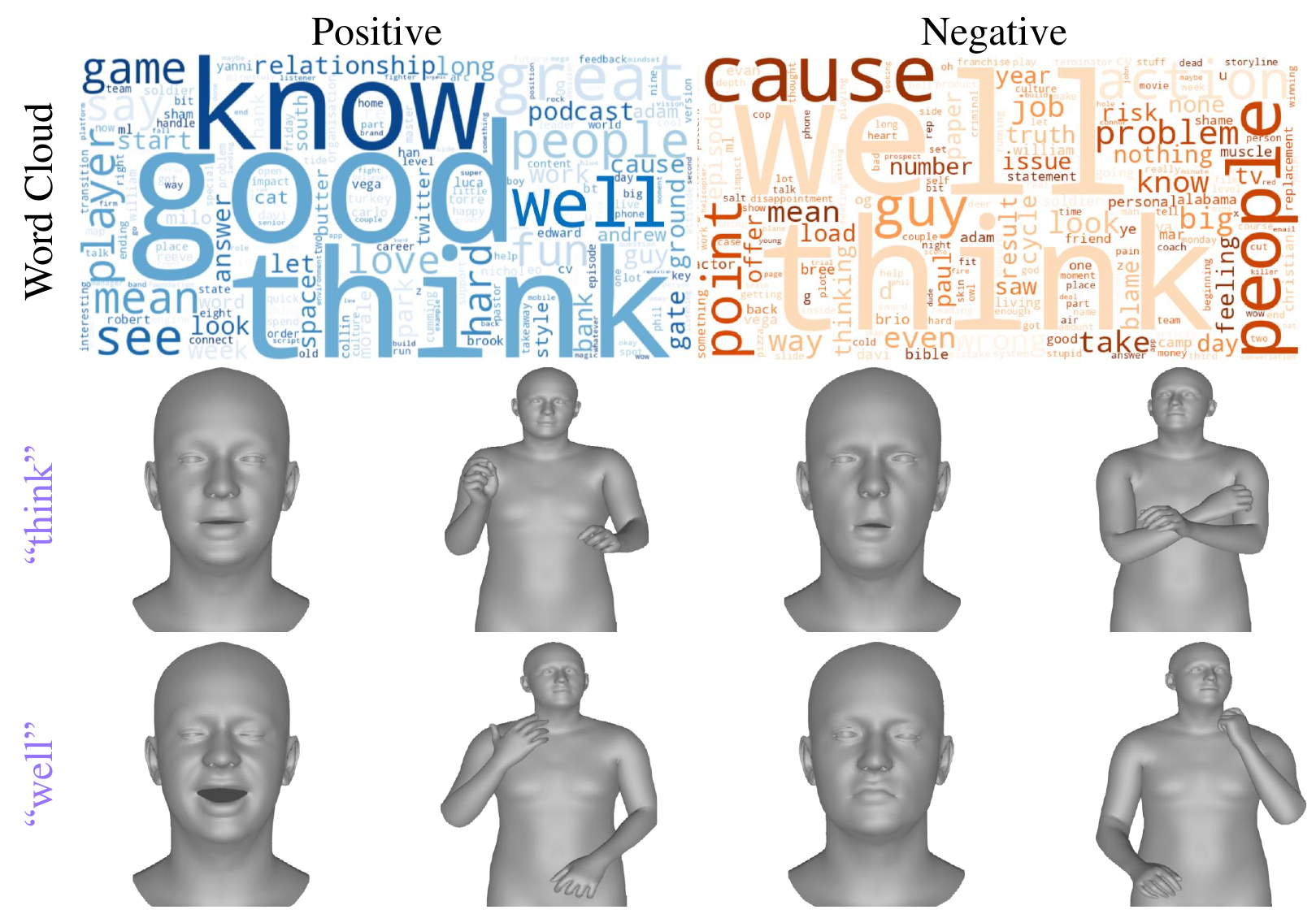}
    \caption{\textbf{Word cloud for text-based sentiment analysis.} It illustrates changes in facial expressions and body language when each word carries a positive or negative context.}
    \label{fig:sent_data_analys}
\end{figure}

These results were visualized using a word cloud, as illustrated in Figure~\ref{fig:sent_data_analys}.
First, an analysis of the words reveals positive and negative associations with certain professions and religions, with ``soldier'' appearing in both positive and negative contexts. Interestingly, in real-world conversations, ``Friday'' is often associated with positive sentiment, while ``Monday'' is linked to negative sentiment.

Also, Figure~\ref{fig:sent_data_analys} shows the nonverbal cues associated with words such as ``think'' and ``well'', comparing their usage in positive versus negative sentiment contexts.
For words like ``think'' and ``well'', sentiments are not prominently reflected in body language.
However, these words often convey a thoughtful or pondering demeanor. 
Notably, facial expressions tend to include frowning when spoken with negative sentiments.
We can infer from these results that nonverbal cues are closely related to sentiment, and leveraging these expressions can enhance the understanding and interpretation of conversations.

\subsection{\DATASETNAME Annotation}
\begin{figure*}
    \centering
    \includegraphics[width=\textwidth]{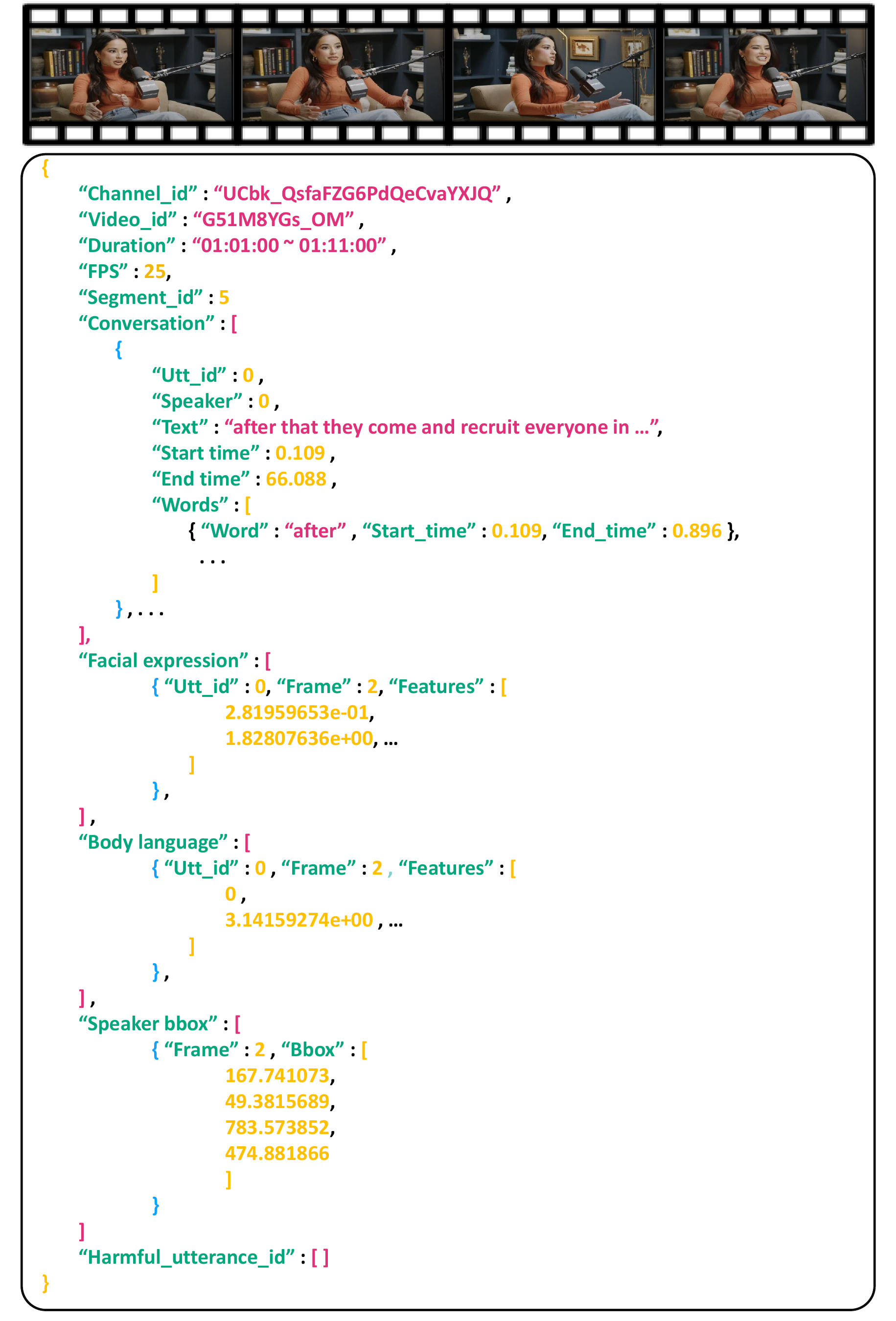}
    \caption{\textbf{\DATASETNAME annotation format.} This is an example of an annotation for a single segmented video. We provide the \DATASETNAME dataset in JSON format. }
    \label{fig:venus-annotation}
\end{figure*}

In this section, we describe the annotation structure of the \DATASETNAME dataset, as illustrated in Figure~\ref{fig:venus-annotation}.

The primary keys in \DATASETNAME include ``Channel ID'', ``Video ID'', ``Duration'', ``FPS'', ``Segment ID'', ``Conversation'', ``Facial expression'', ``Body language'', ``Speaker bbox'' and ``Harmful utterance ID''.
Among these, ``Conversation'' key contains the complete conversation information for a specific video segment, encompassing all data related to utterances.
Within ``Conversation'' key, the ``Words'' key provides time-aligned word information and their corresponding timestamps for each utterance, ensuring temporal alignment of words within the utterance.
``Facial expression'' and ``Body language'' keys represent all nonverbal cue features within the video segment. These nonverbal features are provided alongside utterance IDs and frame information to enable mapping between utterances and features. 
Features of ``Facial expression'' include a total of $153$ features, encompassing information about facial shape, expressions, and jaw.
Meanwhile, features of ``Body language'' comprises $179$ features, which include details about the root of the body, upper and lower body, left and right hands, jaw, and overall body shape.
``Speaker bbox'' represents the results of active speaker detection, providing information about the speaker location in each frame. This information is expressed in the form of coordinates $[x_{\text{top}}, y_{\text{top}}, x_{\text{bottom}}, y_{\text{bottom}}]$, accurately indicating the detected speaker's region in every frame.
Finally, we introduce the ``Harmful utterance ID'' key to mark utterances identified as harmful by our safety strategy. If an utterance ID is included under this key, it does not appear in the ``Conversation'' key. This approach allows us to preserve the maximum amount of video data by retaining all safe utterances while filtering out those deemed harmful, thereby maintaining both ethical standards and dataset integrity.

\subsection{\DATASETNAME Visualization}
\begin{figure*}
    \centering
    \includegraphics[height=0.9\textheight]{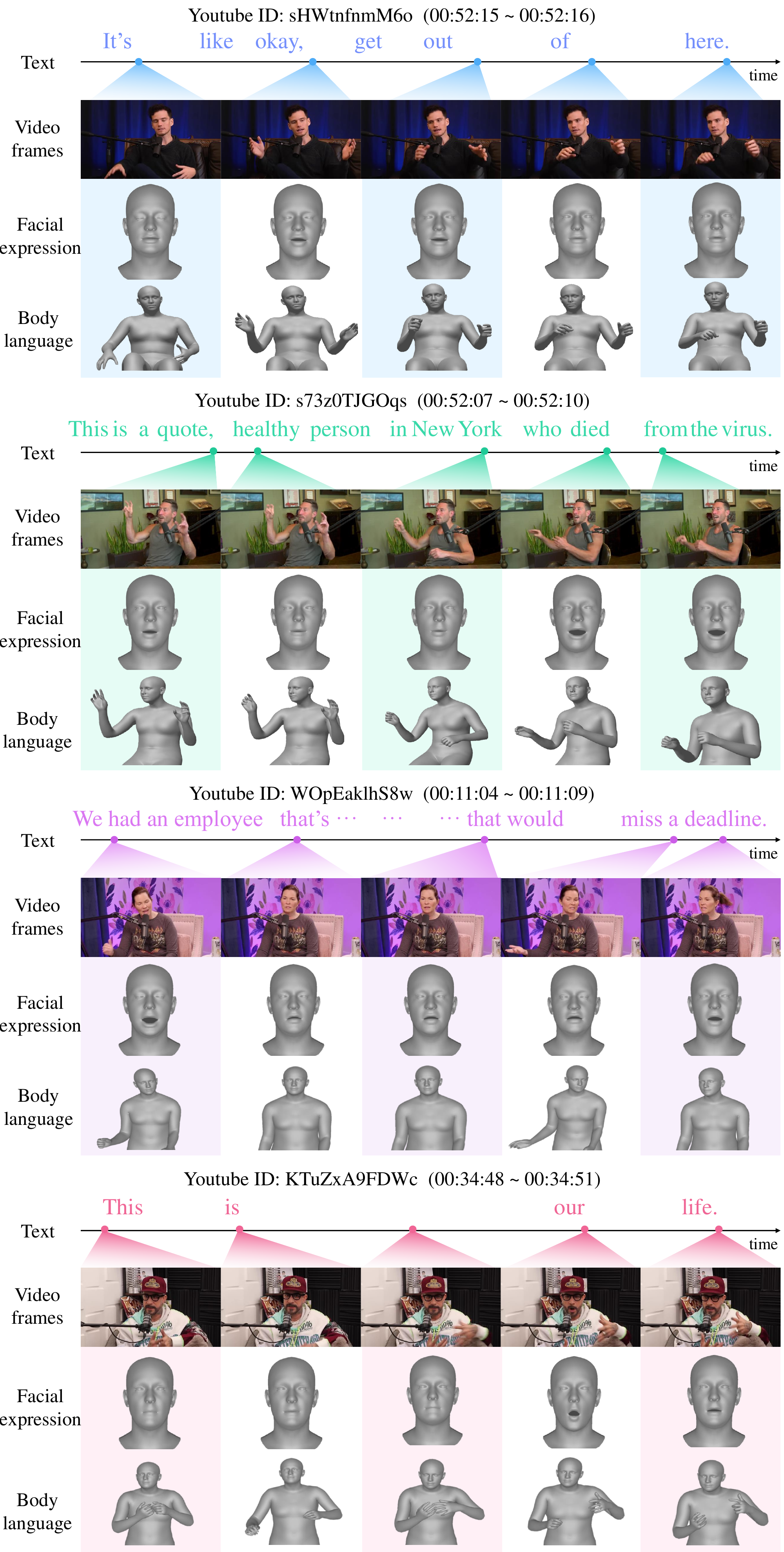}
    \caption{\textbf{Visualization for \DATASETNAME dataset.} This demonstrates the capability of the \DATASETNAME dataset to capture multimodal communication, encompassing speech, body language, and facial expressions. Words are time-aligned using WhisperX, with YouTube IDs providing access to ground truth transcription. ``$\cdots$'' indicates an omission in the text.}
    \label{fig:qual_dataset}
\end{figure*}

We present data visualizations to demonstrate the high quality of the annotated nonverbal expressions in our dataset.
For visualization, we converted the FLAME parameters from EMOCA-v2 to the SMPL-X parameters. As shown in Figure~\ref{fig:qual_dataset}, \DATASETNAME effectively captures key nonverbal expressions, including facial expressions and body language. 

In the first video of Figure~\ref{fig:qual_dataset}, the phrase ``\textit{get out}'' is accompanied by a gesture resembling throwing something away from the speaker. In the second video, the word ``\textit{quote}'' is articulated with a hand gesture resembling air quotes, emphasizing the quoted content in the speech.
These represent the emphasis and intended meaning that nonverbal expressions add to verbal interactions.
\DATASETNAME annotates these expressions, ensuring a rich representation of the subtle, yet essential, aspects of human interaction.

\begin{figure}[t!]
    \centering
    \includegraphics[width=\linewidth]{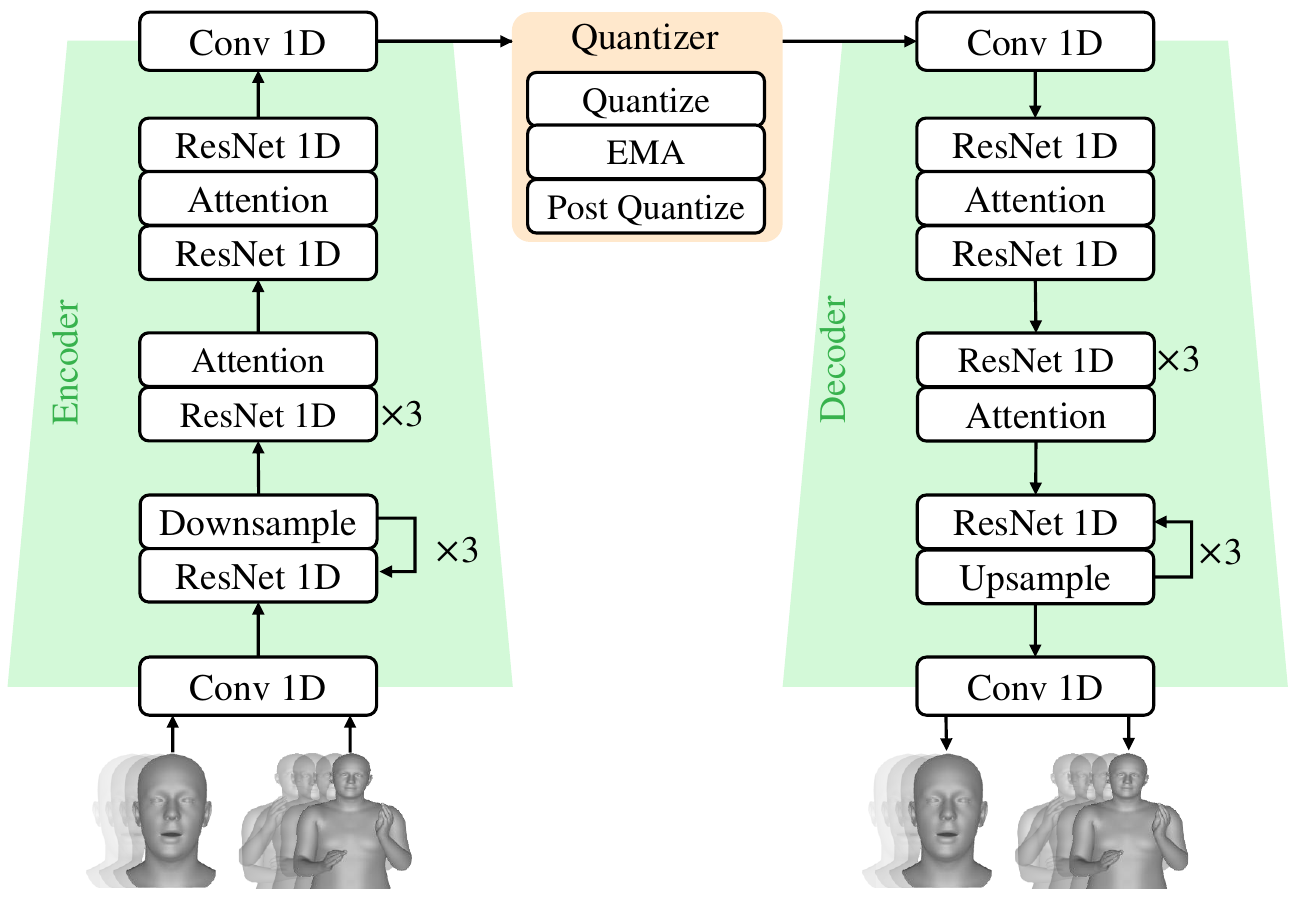}
    \caption{\textbf{Overview of VQ-VAE architecture.} Encoder (left) quantizes the speaker's noverbal-cues, while the decoder (right) projcets the learned discrete codebook tokens back into continuous nonverbal-cues sequence space. The downsampling block consists of 1D convolutional layers with a stride of $2$. Both the Face VQ-VAE and Body VQ-VAE follow the same architecture.}
    \label{fig:vq-vae}
\end{figure}

\section{Details of VQ-VAE}
We trained a VQ-VAE to quantize facial expressions and body language patches, which are utilized as the input and output for the predictor model. Our Face VQ-VAE and Body VQ-VAE were constructed based on the structure proposed by~\cite{guo2024:momask}, with the internal detailed illustrations provided in Figure~\ref{fig:vq-vae}.

\subsection{Implementation Details}
For our VQ-VAE, we use a codebook size of $512$ and set the downsampling factor $q = 8$ in the encoder.
When training, we set the sequence length, $W=512$, to effectively learn utterance-level sequences, with shorter utterances padded with zeros.
The learning rate is initialized at $1e-4$, and the model is trained for $100$ epochs.
We set $10\%$ warmup steps and apply a learning rate decay of 0.1 after $50\%$ steps and $0.01$ after $75\%$ steps. 
For regularization and optimization, we employ EMA with a decay rate of 0.99, L2 regularization with weight decay of 0.1, gradient clipping with a maximum norm of 1.0, and gradient accumulation over 4 steps. We also apply L2 normalization to the codebook vectors.
The optimal model checkpoint is selected based on the validation reconstruction loss.

When codebook learning in $L_{vq}$, we set commitment loss weight, $\beta=0.02$.
For the Face VQ-VAE, the the reconstruction loss weight $\lambda^{f}_{recon}$ is set to $1$, with $\lambda^{\psi}_{recon} = 1$ and $\lambda^{jaw}_{recon} = 5$, determined empirically. And the face velocity loss weight $\lambda^{f}_{vel}$ is set to $0.5$, with $\lambda_{\theta} = 5$ is also empirically chosen.
Similarly, for the Body VQ-VAE, the reconstruction loss weight and velocity loss weight are set to $\lambda^{b}_{recon} = 1$ and $\lambda^{b}_{vel} = 0.5$, respectively.

\subsection{Evaluation Metrics}
\label{appendix:vq_eval_metrics}
To evaluate the performance of the VQ-VAE, we utilize several metrics to assess both realism and diversity.
These evaluation metrics are inspired by prior works~\cite{ng2023:can, zhang2023:generating, liu2024:emage}
We denote ground-truth motion features and generated motion features as $m_{gt}$, and $m_{pred}$.
For realism, we calculate the \textbf{window Vertex L2}, \textbf{VMSE}, and \textbf{LVD} while for diversity, we calculate the \textbf{diversity} and \textbf{variance}.

\noindent \textbf{VMSE}. This metric evaluates the reconstruction error by calculating the mean squared difference between predicted and ground truth vertices in 3D space, offering an intuitive and precise measure of geometric accuracy. We denote the function that maps to the vertex space as $\mathbf{V}(\cdot)$ and the VMSE is defined as follows:
\begin{equation}
    \text{VMSE} = \frac{1}{N} \sum_{i=1}^N||\mathbf{V}(m_{pred,i}) - \mathbf{V}(m_{gt,i})||^{2}_{2}.
\end{equation}
\noindent \textbf{LVD}. This is a metric similar to VMSE, measuring the L1 distance in the vertex space, and it is defined as follows:
\begin{equation}
    \text{LVD} = \frac{1}{N} \sum_{i=1}^N||\mathbf{V}(m_{pred,i}) - \mathbf{V}(m_{gt,i})||_{1}.
\end{equation}
\noindent \textbf{Window Vertex L2}. This metric evaluates the temporal consistency of predicted motion by computing the L2 distance between the averaged ground-truth and predicted vertex positions over sliding windows:
\begin{equation}
    wVL2= \frac{1}{W} \sum_{i=1}^{W} \left\| \frac{1}{S} \sum_{j=1}^{S} \mathbf{V}_{gt}^{(i,j)} - \frac{1}{S} \sum_{j=1}^{S} \mathbf{V}_{pred}^{(i,j)} \right\|_2^2
\end{equation}

\noindent \textbf{Diversity}. This metric quantifies the variability of motion parameters by assessing the spatial distance between selected pairs, providing the diversity of motion representations. 
This follows as:
\begin{equation}
    \text{Diversity} = \frac{1}{K} \sum_{k=1}^{K} \left\| m_{i_k} - m_{j_k} \right\|^2_2,
\end{equation}
where $K$ represents the number of randomly selected pairs, while $m_{i_k}$ and $m_{j_k}$ denote the motion parameters from the first and second indices, respectively.
Here, we randomly selected 1,000 pairs ($K=1,000$) and computed the diversity by repeating this process 10 times.

\noindent \textbf{Variance}. This metric quantifies the average temporal variability of motion parameters.  
Given a motion sequence with \(T\) frames and \(D\) parameters, where \(\mathbf{m}_d \in \mathbb{R}^T\) represents the trajectory of the \(d\)-th parameter over time and \(\bar{\mathbf{m}}_d\) is its mean, the variance is computed as the mean of per-parameter temporal variances:
\begin{equation}
    \text{Variance} =  \frac{1}{D} \sum_{d=1}^D \frac{1}{T} \sum_{t=1}^T (m_{d,t} - \bar{m}_d)^2
\end{equation}

\section{Details of \MODELNAME}
\label{sec:appendix_mars}
\subsection{Details}
We trained \MODELNAME using the LLaMA 3.2-Instruct and Qwen 2.5-Instruct formats and incorporated a system prompt to enhance the model's understanding of nonverbal tokens. 
This is presented in Table~\ref{tab:prompt_eval_winrate}. For supervised fine-tuning,  we set the batch size per GPU at $8$ and the maximum sequence length at $4,096$, and trained over a total of 50 epochs. During inference, we set the maximum sequence length to $512$.

\subsection{Evaluation Metrics}
\textbf{BERT-score}~\cite{zhang2019:bertscore} evaluates the similarity between generated text and reference text at a deeper semantic level. It leverages contextual embeddings derived from pre-trained BERT models to compare candidate and reference tokens. By computing F1 scores based on the cosine similarity of these embeddings, BERTScore provides a nuanced and robust assessment of the semantic alignment and quality of the generated outputs.

\noindent \textbf{Negative Log-Likelihood (NLL)}~\cite{bengio2000:neural} is a function that guides the training of probabilistic models by maximizing the likelihood of the observed data. It measures the discrepancy between the probability distribution predicted by the model and the actual observed data, thereby evaluating how well the model approximates the true data distribution.

\noindent \textbf{PPL}~\cite{bengio2000:neural}, or perplexity, quantifies how effectively a language model predicts the next word in a sequence. Lower perplexity values signify greater confidence and accuracy in the model's predictions, indicating higher quality in generating coherent and contextually appropriate outputs.

\noindent \textbf{METEOR}~\cite{banerjee-lavie-2005-meteor}, short for Metric for Evaluation of Translation with Explicit Ordering, evaluates the quality of generated text by aligning it with the reference text. It incorporates factors like precision, recall, and semantic similarities, such as synonyms and paraphrasing, to provide a more nuanced evaluation.

\begin{table}[ht!]
\centering
\small
\noindent\fbox{
\begin{minipage}{\dimexpr\linewidth-2\fboxsep-2\fboxrule}
\tt
   \textbf{System Prompt}\\
You are a helpful assistant. Text includes nonverbal tokens \texttt{<FACE\_*>}, \texttt{<BODY\_*>} interleaved with language. Help interpret meaning while considering these cues. \\

    \textbf{Input Format} \\
    \{ \\
    \textbf{"role"}: \hspace*{1em} \texttt{["user" / "assistant"]}, \\
    \textbf{"name"}: \hspace*{1em} \texttt{[role\_ID]}, \\
    \textbf{"content"}: \texttt{"Text interleaved with special tokens} \\
    \texttt{<FACE\_TOKEN\_ID> (facial cues),}
    \texttt{<BODY\_TOKEN\_ID> (body languages)."} \\
    \} \\
    
   \textbf{Examples} \\
\{ \\
    \textbf{"role"}: \hspace*{1em}  "user", \\
    \textbf{"name"}: \hspace*{1em} "crXEd-NEsS8\_000\_9" \\
    \textbf{"content"}: "Yeah, \textit{<FACE\_259>}\textit{<BODY\_172>} do you have one of those little \textit{<FACE\_12>} \textit{<BODY\_359>} things in your car?" \\
\} \\

\{ \\
    \textbf{"role"}: \hspace*{1em}  "assistant", \\
    \textbf{"name"}: \hspace*{1em} "crXEd-NEsS8\_000\_10" \\
    \textbf{"content"}: "I have \textit{<FACE\_12>}\textit{<BODY\_239>}\textit{<FACE\_251>}\textit{<BODY\_492>} one." \\
\} \\
\end{minipage}
}
\caption{Input for training \MODELNAME}
\label{tab:prompt_eval_winrate}
\end{table}

\end{document}